\documentclass{article}

\usepackage[numbers]{natbib}
\usepackage[final]{neurips_distshift_2023}

\usepackage[utf8]{inputenc} 
\usepackage[T1]{fontenc}    
\usepackage{hyperref}       
\usepackage{url}            
\usepackage{booktabs}       
\usepackage{amsfonts}       
\usepackage{nicefrac}       
\usepackage{microtype}      
\usepackage{xcolor}         

\usepackage{multirow}
\usepackage{pdfpages}
\usepackage[counterclockwise, figuresleft]{rotating}
\usepackage{amsmath}
\usepackage{todonotes}

\title{An Empirical Study of Uncertainty Estimation Techniques for Detecting Drift in Data Streams}

\author{%
  Anton Winter\\
  TU Darmstadt, Germany\\
  \texttt{anton.winter@stud.tu-darmstadt.de}\\
  \And
  Nicolas Jourdan\\
  TU Darmstadt, Germany\\
  \texttt{n.jourdan@ptw.tu-darmstadt.de}\\
  \And
  Tristan Wirth\\
  TU Darmstadt, Germany\\
  \texttt{tristan.wirth@gris.tu-darmstadt.de}\\
  \And
  Volker Knauthe\\
  TU Darmstadt, Germany\\
  \texttt{volker.knauthe@gris.tu-darmstadt.de}\\
  \And
  Arjan Kuijper\\
  TU Darmstadt, Germany\\
  \texttt{arjan.kuijper@igd.fraunhofer.de}\\
}

\begin{document}

\maketitle


\begin{abstract}
%
In safety-critical domains such as autonomous driving and medical diagnosis, the reliability of machine learning models is crucial. 
One significant challenge to reliability is concept drift, which can cause model deterioration over time. 
Traditionally, drift detectors rely on true labels, which are often scarce and costly. 
This study conducts a comprehensive empirical evaluation of using uncertainty values as substitutes for error rates in detecting drifts, aiming to alleviate the reliance on labeled post-deployment data. 
We examine five uncertainty estimation methods in conjunction with the ADWIN detector across seven real-world datasets. 
Our results reveal that while the SWAG method exhibits superior calibration, the overall accuracy in detecting drifts is not notably impacted by the choice of uncertainty estimation method, with even the most basic method demonstrating competitive performance. 
These findings offer valuable insights into the practical applicability of uncertainty-based drift detection in real-world, safety-critical applications.
\end{abstract}

\section{Introduction and Motivation}

In high-stakes scenarios, such as industrial or medical applications, ensuring the reliability of machine learning model predictions is paramount. 
These domains often present dynamic and uncertain environments, necessitating adaptive machine learning solutions with minimal operational overhead. 
A prevalent issue impacting prediction reliability is \emph{concept drift}, where a data distribution changes over time~\cite{ovadia2019can}. 
It can be denoted as \(P_{\text{{train}}}(X, Y) \ne P_{\text{{online, t}}}(X, Y)\), representing disparities in data distributions during initial training and online operation. 
This is common in various domains, where alterations in conditions lead to non-stationary data streams. 
If overlooked, concept drift can degrade model performance across applications. 
Therefore, adaptive strategies like periodic model updates or retrainings, especially upon drift detection, can be applied to maintain model reliability in evolving operational landscapes. 
However, most conventional drift detection algorithms, e.g. \citep{gama2004learning, baena2006early}, rely on error rates that demand access to scarce and costly true labels. 
An alternative is given by a class of drift detectors that work in an unsupervised way, utilizing a model's prediction confidence / uncertainty as a proxy for the error rate, such as Confidence Distribution Batch Detection (CDBD) \citep{lindstrom2013drift} and Margin Density Drift Detection (MD3) \citep{sethi2015don}. 
More recently, Uncertainty Drift Detection (UDD) was proposed by Baier et al.~\citep{baier2021detecting}, which utilizes neural network uncertainty estimates from Monte Carlo Dropout (MCD) sampling \citep{gal2016dropout} as input for the Adaptive Windowing (ADWIN) detection algorithm~\citep{bifet2007learning}. 
As MCD is only one possibility of extracting uncertainty estimates from neural networks, we investigate the influence of the choice of uncertainty estimation method on the performance of the overall drift detection capability. 
In prior work, Ovadia et al.~\citep{ovadia2019can} analyzed uncertainty estimation methods under dataset shift but only for synthetic drifts. 
While Baier et al.~\citep{baier2021detecting} consider real-world datasets, they limit their experiments to a single uncertainty estimation method. 
Thus, our core contribution is an empiric comparison of four state-of-the-art neural network uncertainty estimation methods, as well as a baseline method, for classification tasks in combination with the ADWIN detector to identify drifts in real-world data streams. 
These uncertainty estimators are evaluated using seven commonly used real-world datsets.\\ 

\section{Methodology and Experiments}
\begin{figure}[h!]
  \centering
  
  \includegraphics[width=12cm]{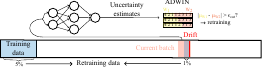}
  
  \caption{Approach to uncertainty drift detection.}
  
  \label{fig:approach}
\end{figure}

To compare the uncertainty estimation methods introduced in the following, we conduct two experiments for each method and dataset. 
Both start by training the method with the initial five percent of the whole data stream.
The first experiment serves as a baseline and thus, the remaining data is tested without analyzing uncertainty estimates or triggering retrainings.
In the main experiment however, batches of the stream are evaluated and uncertainty estimates are used as a proxy for the error rate of the ADWIN detector.
Once a drift is detected, a retraining is triggered with the initial five percent plus the most recent samples equivalent to one percent of the stream size.
Thereby, models may adapt to new concepts while retaining sufficient generalization.
Every experiment is repeated five times with different random seeds and results are averaged to allow for a fair comparison.
Figure \ref{fig:approach} illustrates the process of the main experiment. 

\subsection{Uncertainty Estimation Methods}
\label{uncertainty methods}
To quantify model uncertainty, Bayesian neural networks, which learn a posterior distribution over model parameters, can be employed ~\citep{ovadia2019can}.
This distribution enables the application of Bayesian model averaging (BMA) during inference.
Therefore, multiple weights $w_i$ are drawn to gather a distribution of predictions $p_i(y|w_i,x)$, given input features $x$ and target labels $y$.
The final prediction $\hat{p}(y|x)$ is then given as the average

\begin{equation}
    {\hat{p}(y|x) = \frac{1}{P} \sum_{i=1}^{P}p_i(y|w_i,x).}
\end{equation}

For regression tasks, the uncertainty is the standard deviation of said distribution.
While there are several uncertainty-related metrics for classification tasks, only Shannon's entropy $H$ does not require ground-truth labels.
Given the final prediction $\hat{p}(y|x)$ with $K$ classes, it is computed as 

\begin{equation}
    {H[\hat{p}(y|x)] = -\sum_{k=1}^{K}\hat{p}(y=k|x) \cdot \textrm{log}_2\hat{p}(y=k|x).}
\end{equation}

Although bayesian methods were previously considered state-of-the-art, they are computationally intractable for modern neural networks with millions of parameters ~\citep{maddox2019simple}.
Therefore, alternatives have been developed, of which we analyzed the following in our experiments. 
To get an uncertainty estimate, Shannon's entropy $H$ is applied to the final prediction of each method.

\textbf{Basic Neural Network.}
Given the focus on classification tasks, a distribution of predictions is not necessarily required.
Hence, the simplest method is to use a single prediction from an unmodified neural network.
The motivation for this is to have a baseline for the more sophisticated methods. 

\textbf{Monte Carlo Dropout (MCD).}
Rather than drawing multiple weights from a posterior distribution as in BMA, a random dropout filter is applied to the neurons for several forward passes.
These estimates are then averaged to get a final prediction.
This allows for estimating the uncertainty in the model parameters based on the variability of the predictions across different dropout masks ~\citep{baier2021detecting,gal2016dropout}.

\textbf{Ensemble.} 
A distribution of predictions can also be won by training multiple neural networks.
Different seeds of members introduce randomness due to their influence on the initial weights as well as the shuffling of data during training.
As Lakshminarayanan et al.~\citep{lakshminarayanan2017simple} have shown, few members, i.e. 5, can be sufficient for good uncertainty estimates.

\textbf{Stochastic Weight Averaging Gaussian (SWAG).}
Based on Stochastic Weight Averaging (SWA), a generalization technique in deep networks, Maddox et al.~\citep{maddox2019simple} propose a method to approximate a posterior distribution over neural network weights.
Therefore, a Gaussian is fit utilizing the SWA solution as the first moment and a low rank plus diagonal covariance also inferred from stochastic gradient descent iterates.
Given this posterior distribution, BMA is applied to get a final prediction.

\textbf{Activation Shaping (ASH).}
The ASH method can be considered a more advanced version of the basic neural network, as it also works on single predictions.
Djurisic et al.~\citep{djurisic2022extremely} introduced it as an out-of-distribution (OOD) detection method that reaches state-of-the-art performance.
Assuming over-parameterized feature representations in modern neural networks, the hypothesis is that pruning a larger percentage of activations in a late layer helps with tasks such as OOD detection.

The hyperparameters of the introduced methods as well as the model architectures can be found in Appendix \ref{reproducability}.
Furtheremore, we include details of the tuning process in Appendix \ref{tuning}.

\subsection{Drift Detector}
\label{sec:detector}
Concept drift detectors, such as Drift Detection Method ~\citep{gama2004learning}, Page Hinkley Test ~\citep{page1954continuous}, and ADWIN ~\citep{bifet2007learning}, are typically error rate-based, necessitating access to costly true labels ~\citep{gonccalves2014comparative}.
In contrast, data distribution-based detectors exclusively analyze input features, often using distance metrics like the Kolmogorov-Smirnov test ~\citep{raab2020reactive} to identify changes in feature distribution.
Regardless of the detection method employed, distinguishing between noise and genuine concept drift poses a significant challenge ~\citep{tsymbal2004problem}, requiring a balance between swift adaptation to changes and resilience to noise. 
ADWIN offers performance guarantees for false positives and false negatives, making it an attractive choice. 
Furthermore, it is able to work with any real-valued input instead of beeing limited to an error rate between 0-1.
As introduced by Bifet et al. ~\citep{bifet2007learning}, ADWIN utilizes sliding windows of variable size. While no drift is present, new samples are added to a window W. After each sample, the algorithm attempts to find two sub-windows $W_0$ and $W_1$ that contain distinct averages. Once this happens a drift is assumed and the older sub-window is discarded.
The variability of heterogeneous real-world data streams can be addressed by the sensitivity parameter $\delta$ $\in$ (0, 1).
The configuration for our experiments can be found in Appendix \ref{reproducability}.

\subsection{Datasets}
For our studies, we use seven real-world classification datasets from the USP Data Stream Repository~\cite{souza2020challenges}.
They encompass abrupt, incremental and reocurring drifts, along with combinations thereof.
In the \textbf{Gas} sensor dataset chemical sensor data is analyzed to identify one of six gases.
The \textbf{Electricity} dataset focuses on predicting market price changes driven by supply and demand.
For the \textbf{Rialto} dataset, segments of images from a timelapse with changing weather conditions shall be classified.
Lastly, optical sensors are used to analyze moving patterns of flying insect species while drift is artificially introduced to generate the \textbf{InsAbr}, \textbf{InsInc}, \textbf{InsIncAbr} and \textbf{InsIncReo} datasets.

\subsection{Metrics and Results}
\label{exp setting}
For evaluation, we focus on the following two metrics to capture the quality of the uncertainty estimates as well as the drift detection performance:
\textbf{Expected Calibration Error (ECE) $\downarrow$}~\citep{naeini} measures the average deviation between prediction confidence and accuracy. As the name suggests, it quantifies how well a model is calibrated. We expect that calibration correlates positively with drift detection capability.
\textbf{Matthew's Correlation Coefficient (MCC)~$\uparrow$} is able to handle class imbalances which generally makes it a good metric for classification tasks \citep{chicco2020advantages}. We employ the MCC to measure the overall prediction performance of the models, averaged over the complete experiment runs. We expect that poor drift detection performance will lead to unsuitable retraining points, in turn producing low MCC scores and vice versa.

\begin{figure}[h!]
  \centering
  
  \includegraphics[width=12cm]{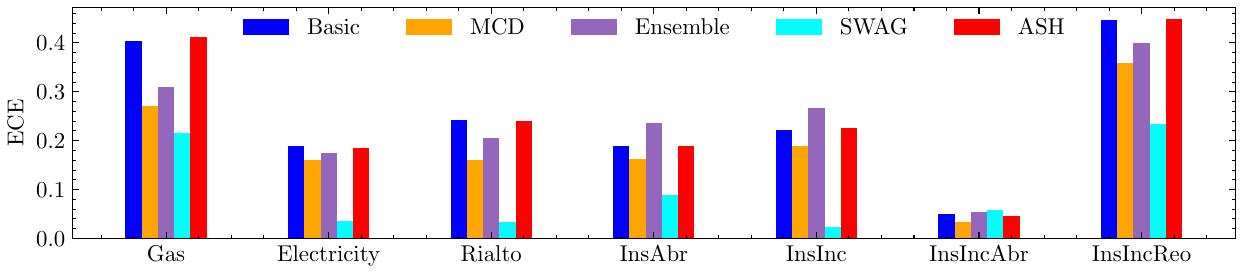}

  \caption{Calibration of the employed uncertainty estimation methods measured by ECE ($\downarrow$) across the seven datasets.}
  \label{fig:ECE}
\end{figure}

The results of our experiments can be found in Table \ref{table:final_comparison}.
Analyzing the MCC values shows that the SWAG method offers the most balanced performance across all datasets.
However, the gap in performance to the other methods is minimal.
In fact, all methods perform fairly similarly. 
Surprisingly, even the basic method without any modifications keeps up with the others. 
%
%
Greater differences can be identified when analyzing the ECE as depicted in Figure \ref{fig:ECE}.
Here, the SWAG method offers significantly better calibrated predictions in nearly all datasets.
The only exception is the InsIncAbr dataset, where all methods achieve a proficient calibration. 
All other methods appear to be similarly worse calibrated compared to SWAG for the remaining datasets.
Despite that, this does not directly translate to a better drift detection performance, as shown by the MCC values.\\
Meanwhile, the total execution time fluctuates notably depending on the method selected, as presented in the last row of Table \ref{table:final_comparison}. 
As the basic and ASH method are based on a single sample, they serve as a lower bound in this regard.
While MCD and the SWAG method both increase the inference runtime due to the sampling process, adaptations in the training process of the SWAG method incurr additional overhead.
Although the execution time of the ensemble could be reduced by parallelizing the training and inference process of individual ensemble members, this would require additional computational resources.
Hence, we choose not to, resulting in the highest execution time by far.
\\
Appendix \ref{additional results} includes further details of the main experiment as well as an additional experiment to validate the retraining positions found by the uncertainty-based detector.
Furthermore, it contains the standard deviations of our experiments.

\begin{table}[htb]
   \centering
   \footnotesize
   
    \caption{Performance comparison of uncertainty estimation methods for drift detection. Table cells contain the average MCC ($\uparrow$) values and (number of retrainings) for the naive baselines without retrainings (upper) and retrainings triggered by ADWIN when using the respective uncertainty estimation method (lower), respectively. Bold numbers indicate the best performance. To given an impression of the computational cost, the last row contains the total execution times for each method.}
    
   \label{table:final_comparison}
    \begin{tabular}{cccccc}
      \toprule
        & Basic & MCD & Ensemble & SWAG & ASH\\
       \midrule
       \multirow{2}{*}{Gas} & 0.273 (0) & 0.256 (0) & 0.245 (0)  & \textbf{0.299} (0) & 0.275 (0)\\
       & 0.455 (36) & 0.46 (55) & \textbf{0.492} (50) & 0.46 (52) & 0.459 (35)\\
       \midrule
       \multirow{2}{*}{Electricity} & 0.178 (0) & \textbf{0.198} (0)& 0.183 (0)& 0.191 (0) & 0.175 (0)\\
       & 0.424 (11) & 0.421 (10) & 0.405 (10) & 0.419 (7) & \textbf{0.438} (10)\\
       \midrule
       \multirow{2}{*}{Rialto} & 0.532 (0) & \textbf{0.534} (0)& 0.505 (0) & 0.52 (0) & 0.525 (0)\\
       & 0.537 (43) & \textbf{0.553} (48) & 0.527 (45) & 0.54 (52) & 0.539 (43)\\
       \midrule
       \multirow{2}{*}{InsAbr} & 0.471 (0)& 0.472 (0)& 0.461 (0)& \textbf{0.48} (0) & 0.474 (0)\\
       & \textbf{0.519} (9) & 0.509 (8) & 0.503 (8) & 0.514 (6) & 0.508 (7)\\
       \midrule
       \multirow{2}{*}{InsInc} & 0.087 (0) & \textbf{0.1} (0)& 0.081 (0)& \textbf{0.1} (0) & 0.085 (0)\\
       & 0.241 (3) & 0.238 (3) & 0.241 (3) & \textbf{0.301} (4) & 0.231 (3)\\
       \midrule
       \multirow{2}{*}{InsIncAbr} & 0.304 (0)& 0.307 (0)& 0.308 (0) & 0.299 (0) & \textbf{0.316} (0)\\
       & 0.53 (24) & 0.525 (26) & 0.518 (23) & 0.445 (25) & \textbf{0.531} (25)\\
       \midrule
       \multirow{2}{*}{InsIncReo}  & 0.141 (0)& 0.133 (0) & \textbf{0.172} (0) & 0.16 (0) & 0.133 (0)\\
       & 0.253 (18) & 0.247 (20) & 0.236 (18) & \textbf{0.302} (21) & 0.243 (20)\\
       \midrule
       Total exec. time & 6821s & 7339s & 15653s & 9036s & 6890s \\
       \bottomrule
       
   \end{tabular}  
\end{table}
\clearpage

%

\section{Conclusion}
In this work, we implemented five uncertainty estimation methods for classification tasks and evaluated them in experiments including seven real-world datasets. 
Our goal was to compare the utility of their uncertainty estimates for unsupervised concept drift detection by using them as a proxy for the error-rate in combination with the ADWIN detector. 
Thereby, drift points in data streams shall be identified to trigger retrainings at the appropriate time and ultimately prevent model decay.
Interestingly, even our baseline method, relying solely on the entropy calculated from the softmax scores, performed competitively with more sophisticated state-of-the-art methods.
Moreover, all methods performed fairly similar in terms of overall classification performance as measured by the MCC metric. 
While the SWAG method achieved the most balanced MCC values, differences were only marginal.
However, this was not the case when analyzing the ECE. 
Here the SWAG method offers significantly better calibrated predictions than all other methods. 
Regardless, these did not translate to better results for the drift detection.
Thus, the assumption can be made, that the choice of method does not have a noteworthy influence on the performance of uncertainty-based concept drift detection for real-world applications.
\\To confirm the previous assumption, future work may include testing further real-world datasets, including regression problems.
For those, the basic neural network and the ASH method are no longer applicable. 
Instead, the effect of the ASH method in combination with the remaining approaches could be studied.


\bibliographystyle{plain}

{\small
\bibliography{bibliography}}

\clearpage

\appendix
\section{Appendix}
\subsection{Reproducibility}
\label{reproducability}
To make our experiments reproducible, Table \ref{table:architectures_overview} gives an overview of the neural network architecture used for each dataset. 
Hidden layers use Rectified Linear Unit activations, while softmax is applied in the final layer.
The ADAM optimizer is used with binary or categorical cross-entropy loss, depending on the number of classes. 
For \textbf{MCD}, 100 forward passes are carried out.
The \textbf{Ensemble} consists of three members. 
Bayesian model averaging is conducted with 100 samples from the posterior approximation of the \textbf{SWAG} method. Details on these choices are discussed in \ref{tuning}.
Furthermore, the estimated covariance matrix utilized in the approach has a rank of 25 and is updated each epoch, starting at the first iteration.
For the \textbf{ASH} method, the version termed ASH-p was chosen, where unpruned activations are not modified at all.
Pruning is applied in the penultimate hidden layer (i.e. third last overall layer) with a pruning percentage of 60\%.
Lastly, Table \ref{table:adwin_deltas} indicates the sensitivity values $\delta$ for the ADWIN detector.

\begin{table}[h!]
   \centering
   
   \caption{Overview of model architectures.}
    \label{table:architectures_overview}
    
    \begin{tabular}{ccccc}
      \toprule
       Name & No. Layers & Neurons per layer & Dropout rate & Epochs\\
       \midrule
       Gas & 5 & 128, 64, 32, 16, 8 & 0.2 & 100\\
       \midrule
       Electricity & 3 & 32, 16, 8 & 0.1 & 400\\
       \midrule
       Rialto & 4 & 512, 512, 256, 32 & 0.2 & 200\\                                         
       \midrule
       InsAbr & 5 & 128, 64, 32, 16, 8 & 0.1 & 200\\        
       \midrule
       InsInc & 5 & 128, 64, 32, 16, 8 & 0.1 & 100\\         
       \midrule
       InsIncAbr & 3 & 32, 16, 8 & 0.1 & 50\\               
       \midrule
       InsIncReo  & 3 & 128, 64, 32 & 0.1 & 400\\      
       \bottomrule                               
    \end{tabular}  
\end{table}

\begin{table}[h!]
    \centering
    
    \caption{Sensitivity values for ADWIN detector.}
    \label{table:adwin_deltas}
     
     \begin{tabular}{ccccccc}
        \toprule
        Gas & Electricity & Rialto & InsAbr & InsInc & InsIncAbr & InsIncReo\\
        \midrule
        0.1 & 1e-15 & 1e-20 & 0.002 & 0.002 & 0.1 & 0.1\\
        \bottomrule
     \end{tabular}  
 \end{table}


\subsection{Additional Results and Experiments}
\label{additional results}
\begin{sidewaysfigure}[!htbp]
   \centering
   \setlength{\tabcolsep}{-2pt}
   
   \vspace{10mm}
   \begin{tabular}{ccccc}
      \hspace{-3mm}
      \includegraphics[width=45mm]{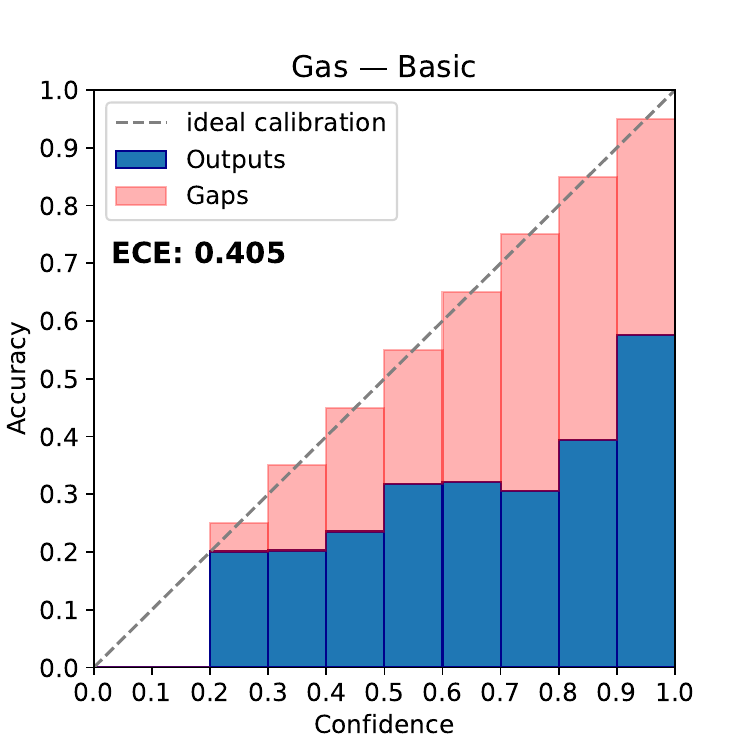}&\includegraphics[width=45mm]{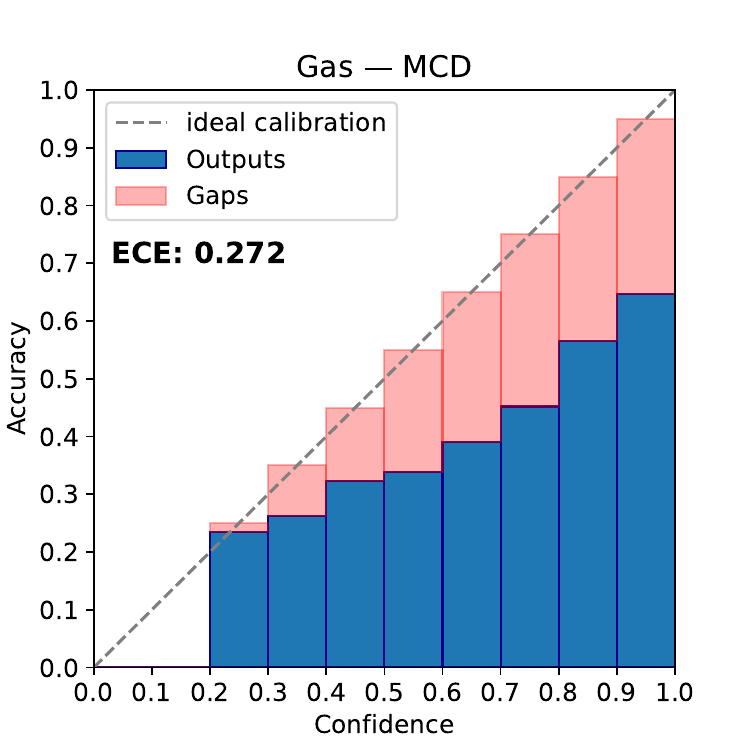}&\includegraphics[width=45mm]{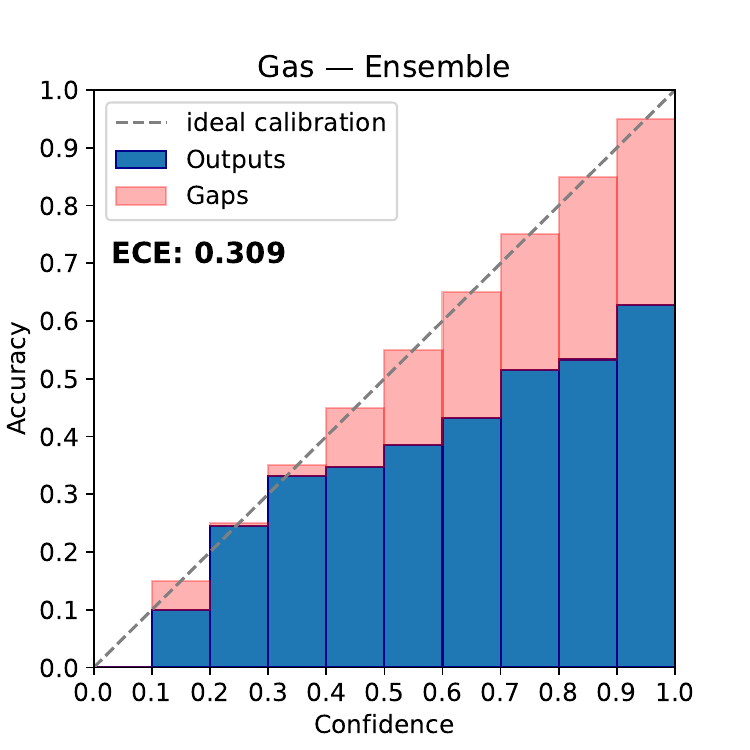}&\includegraphics[width=45mm]{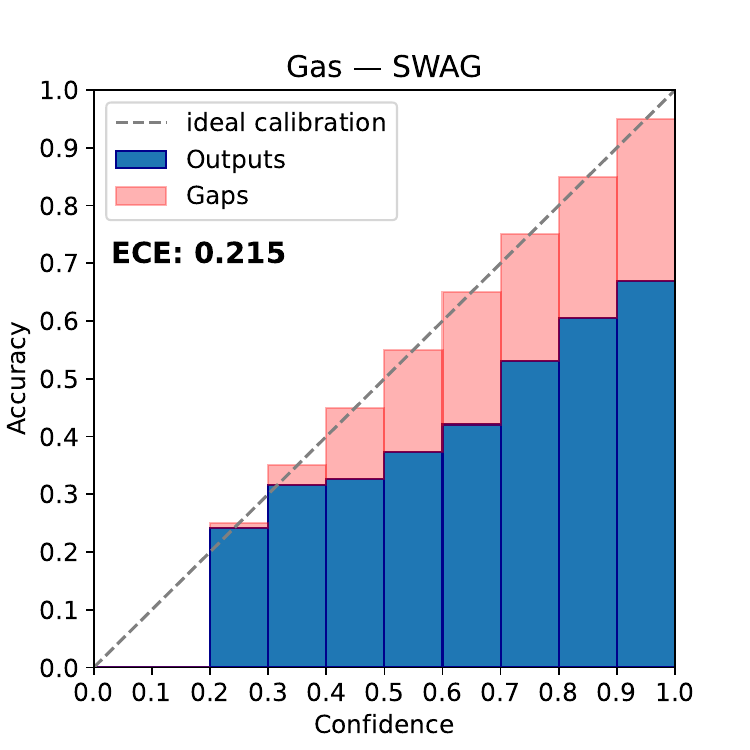}&\includegraphics[width=45mm]{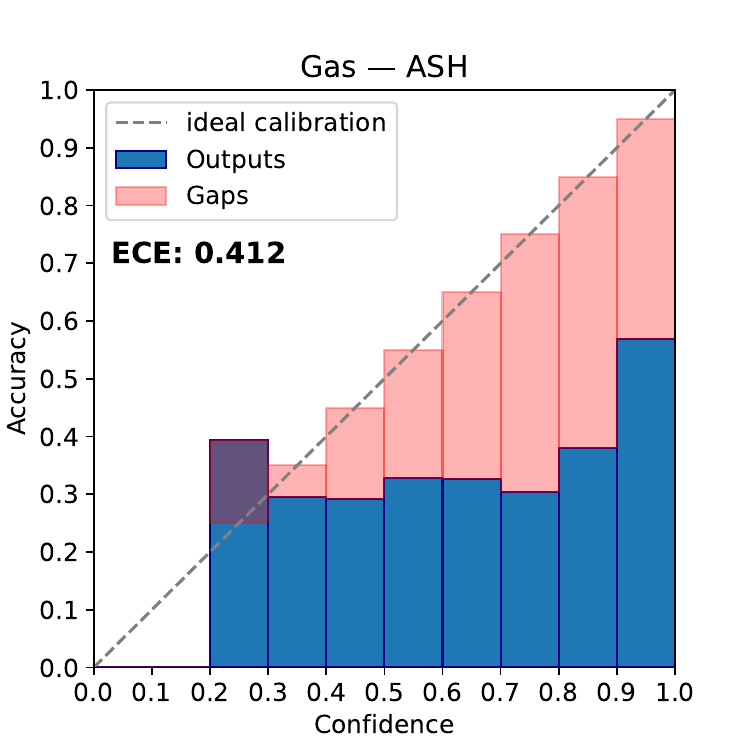}\\
      
      \hspace{-3mm}
      \includegraphics[width=45mm]{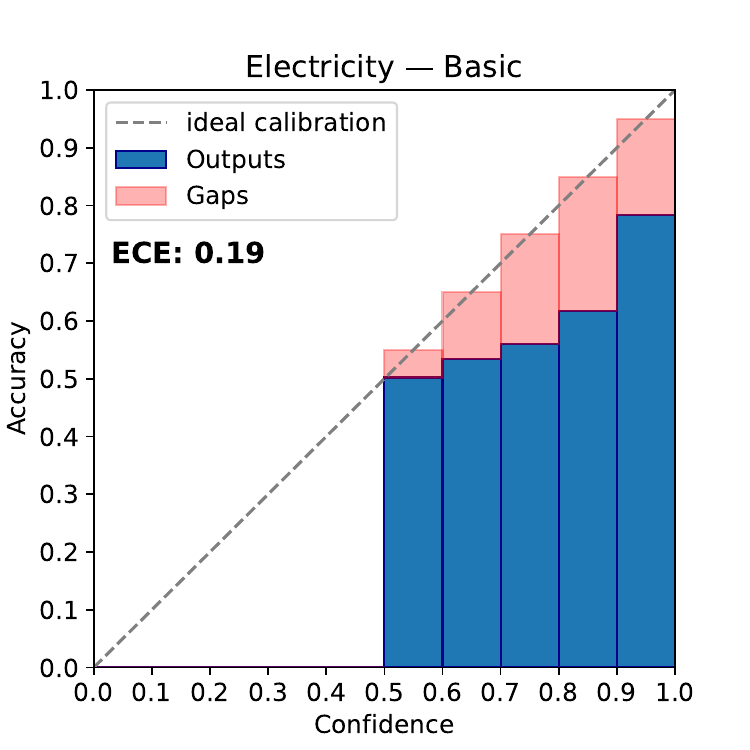}&\includegraphics[width=45mm]{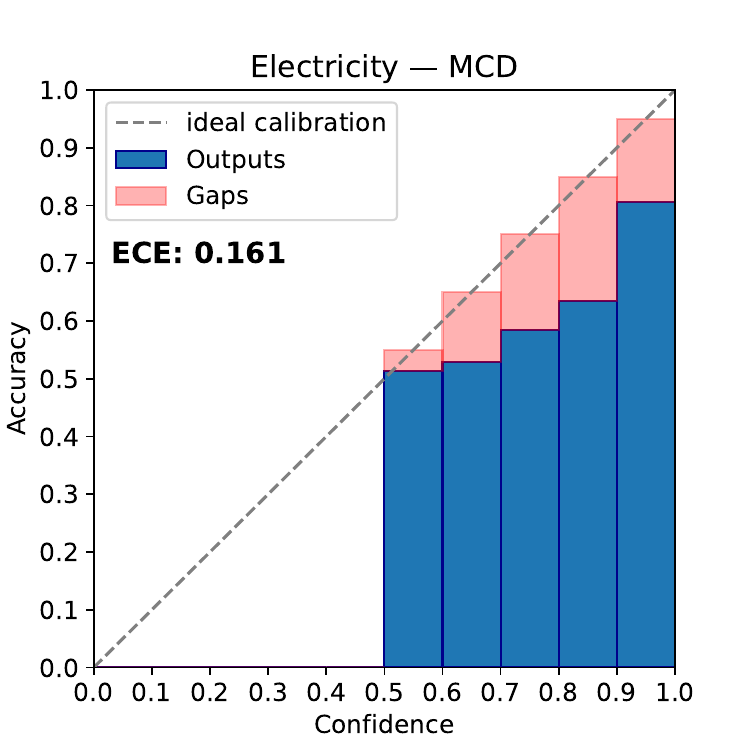}&\includegraphics[width=45mm]{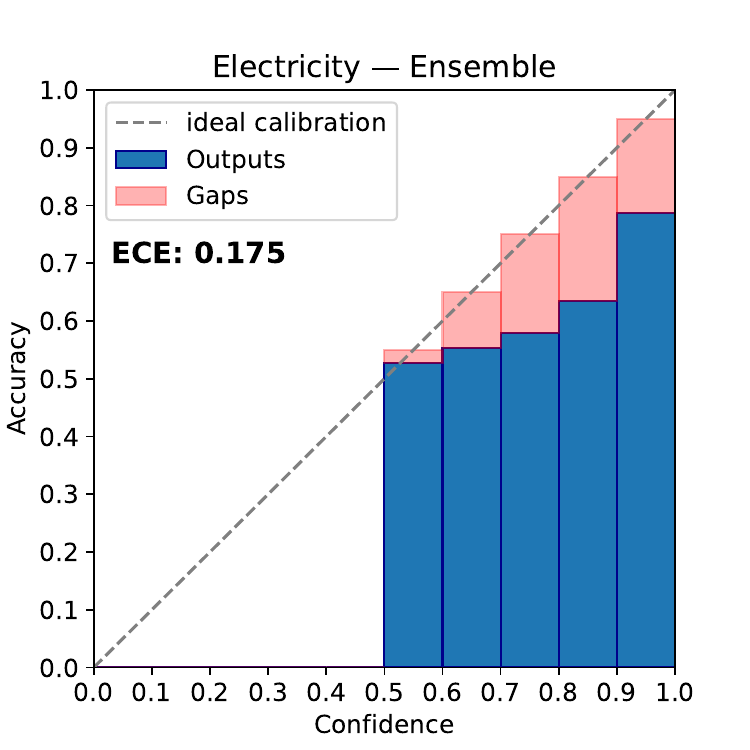}&\includegraphics[width=45mm]{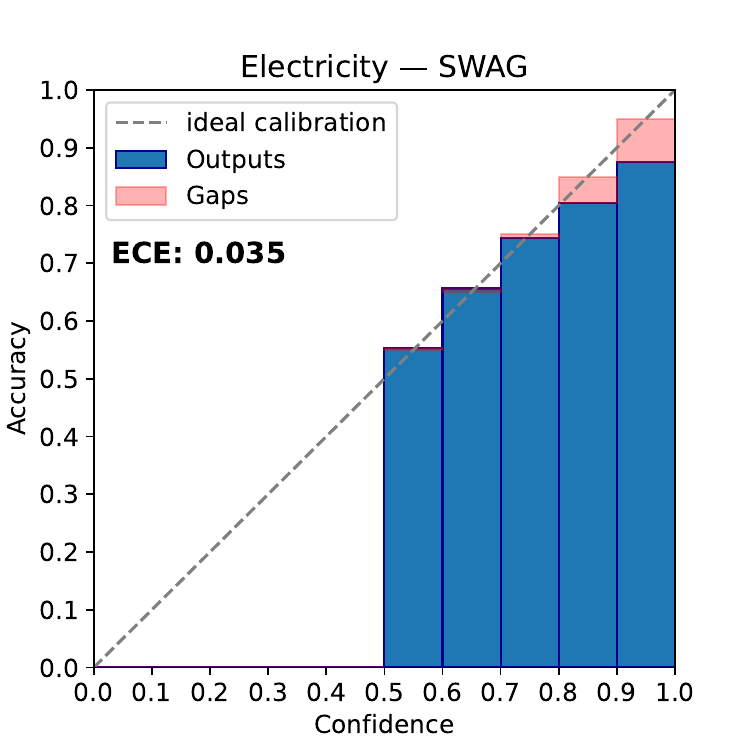}&\includegraphics[width=45mm]{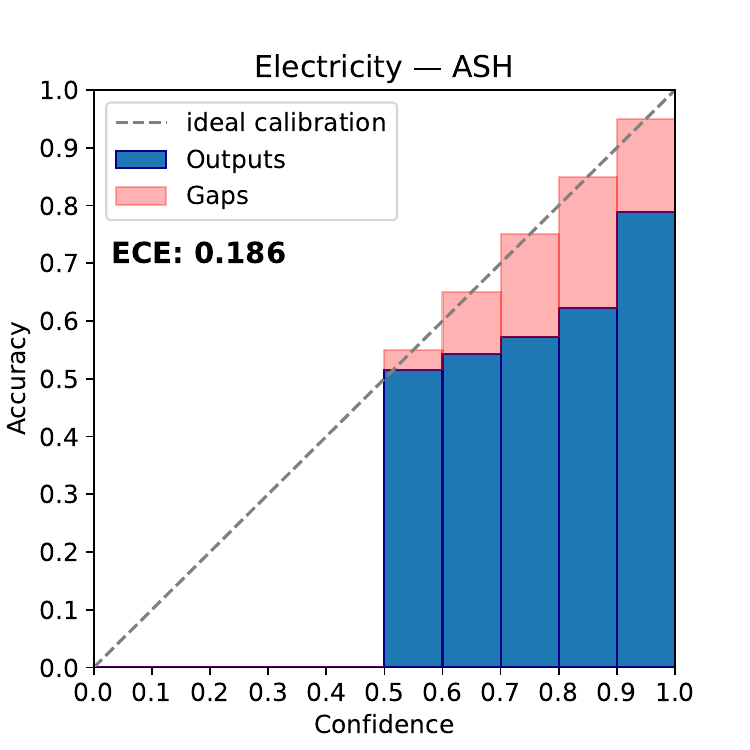}\\
     
      \hspace{-3mm}
       \includegraphics[width=45mm]{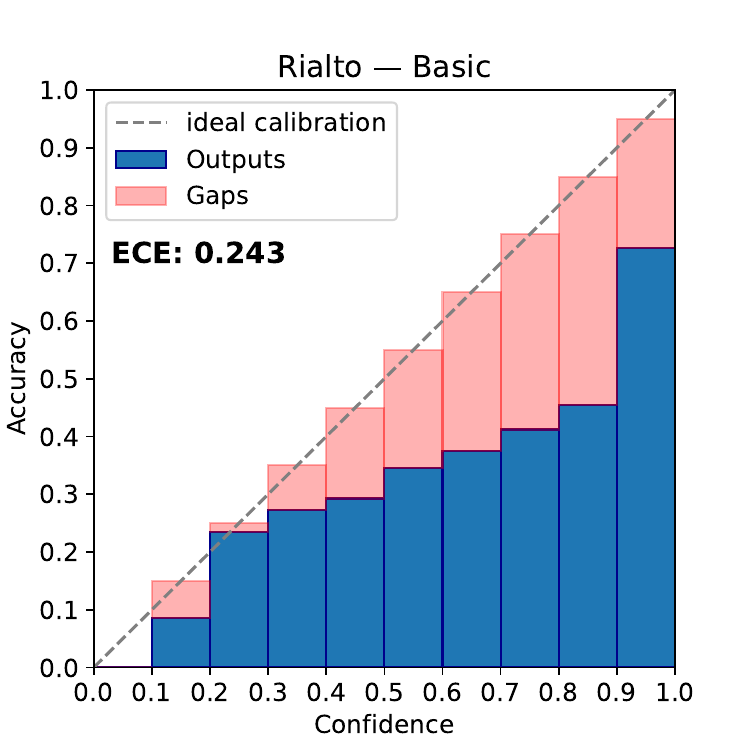}&\includegraphics[width=45mm]{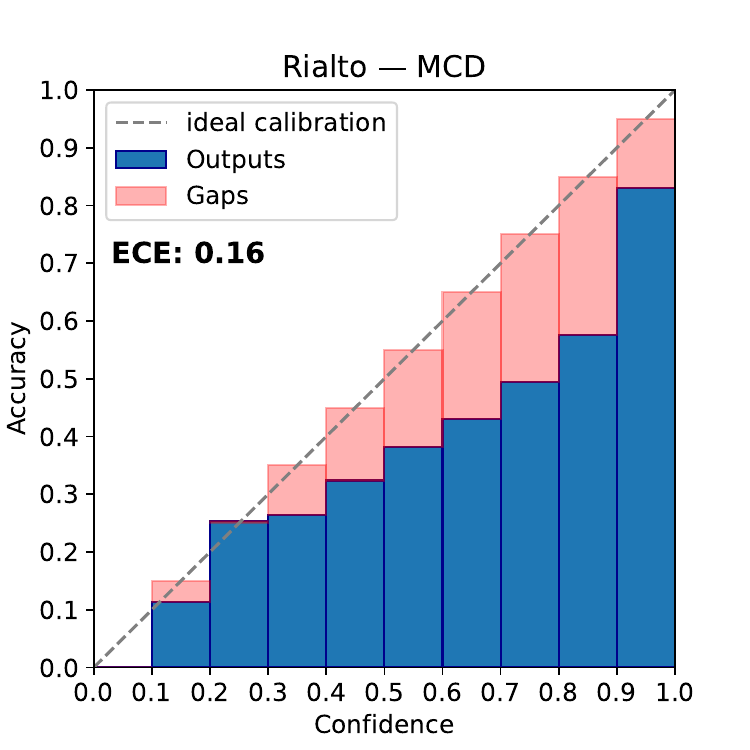}&\includegraphics[width=45mm]{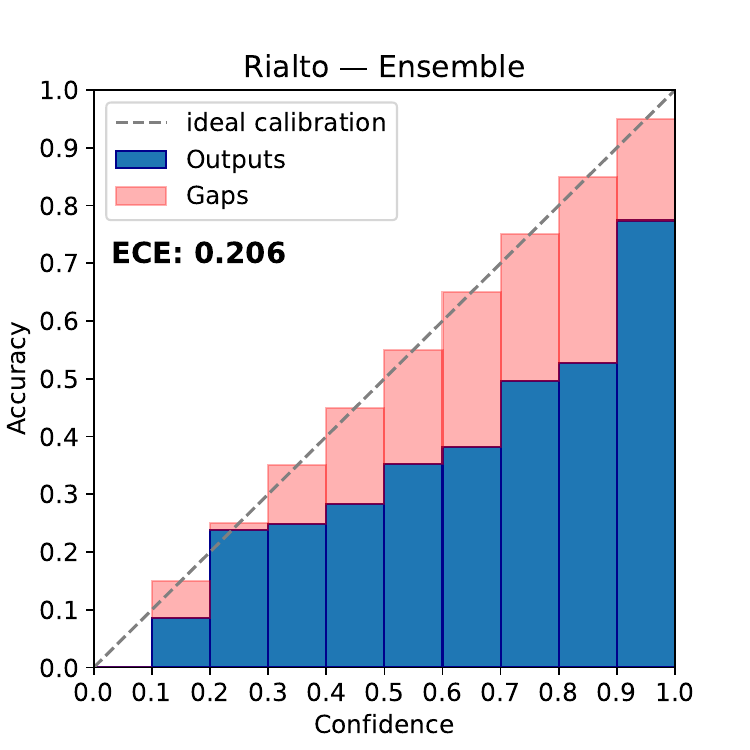}&\includegraphics[width=45mm]{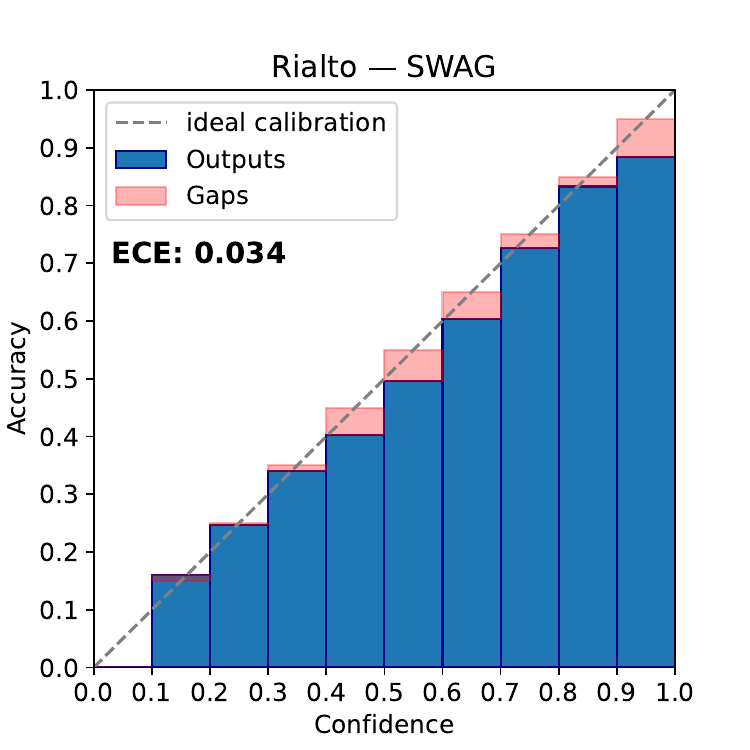}&\includegraphics[width=45mm]{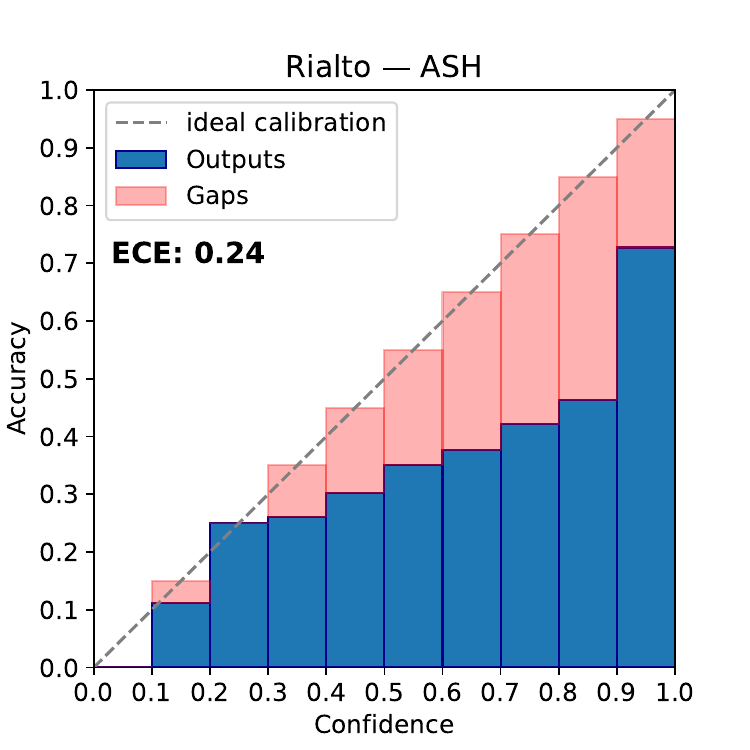}\\
       
   \end{tabular}
   \caption{Reliability diagrams of main experiment (1/3)}
   \label{fig:reliability_1}
\end{sidewaysfigure}

\begin{sidewaysfigure}[!htbp]
      \centering
      \setlength{\tabcolsep}{-2pt}
      
   \vspace{10mm}
      \begin{tabular}{ccccc}

         \hspace{-3mm}
         \includegraphics[width=45mm]{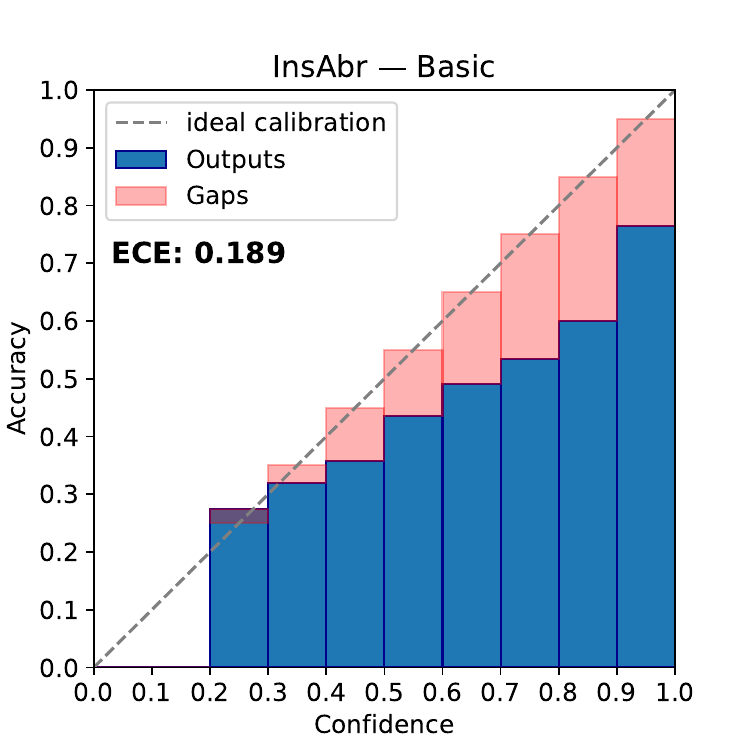}&\includegraphics[width=45mm]{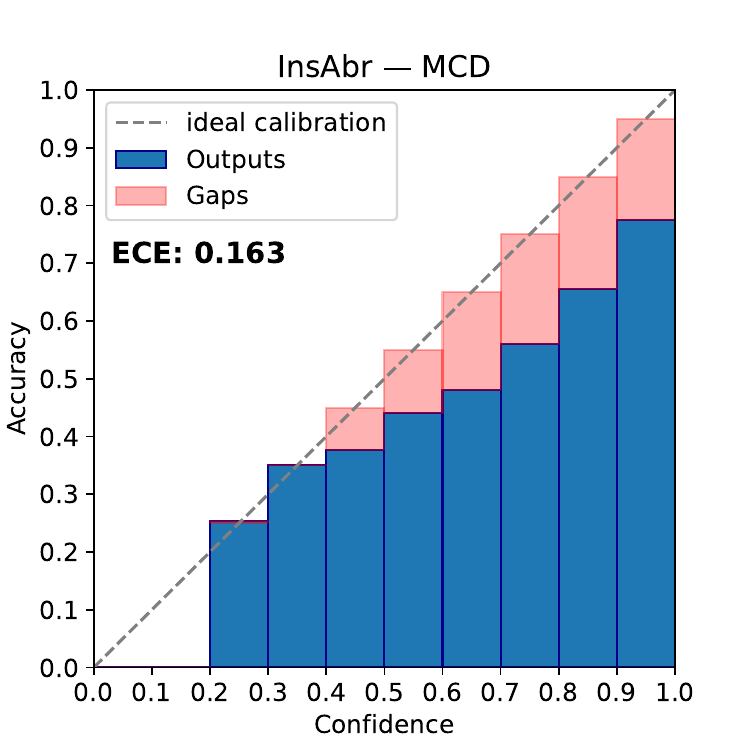}&\includegraphics[width=45mm]{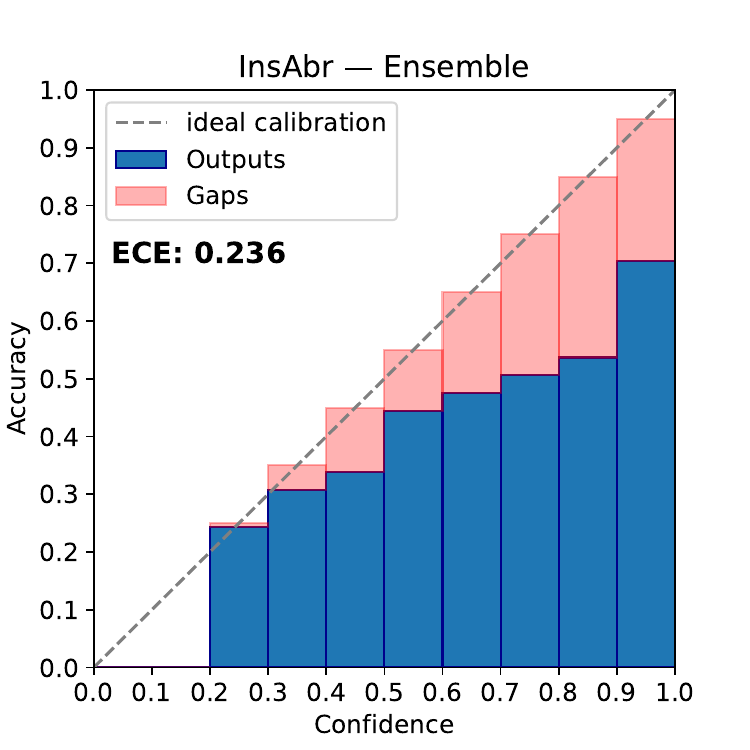}&\includegraphics[width=45mm]{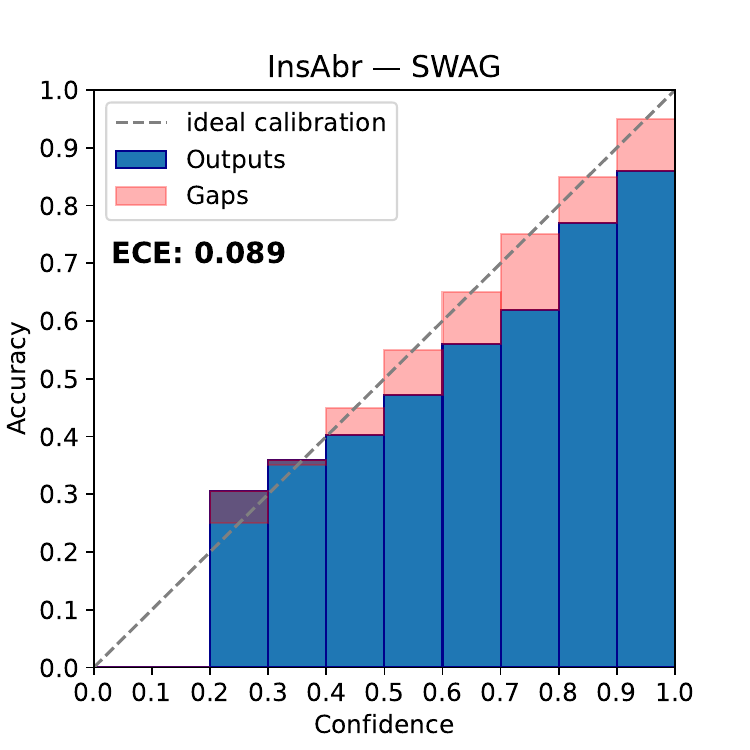}&\includegraphics[width=45mm]{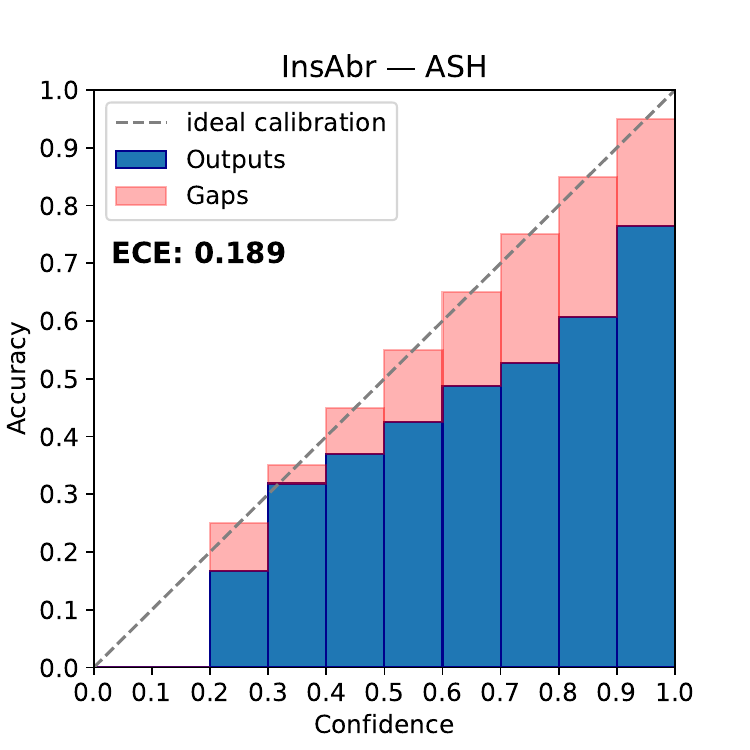}\\
         \hspace{-3mm}
         \includegraphics[width=45mm]{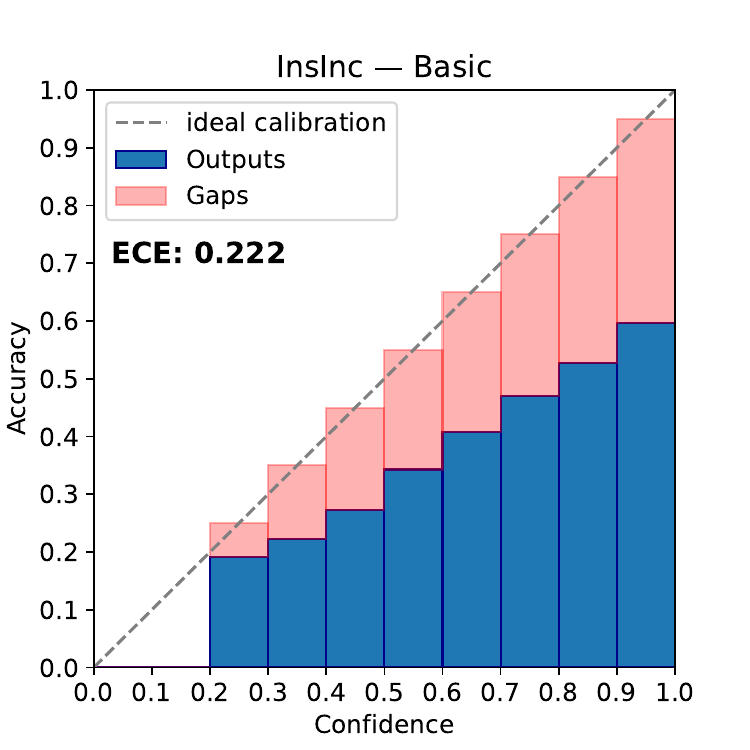}&\includegraphics[width=45mm]{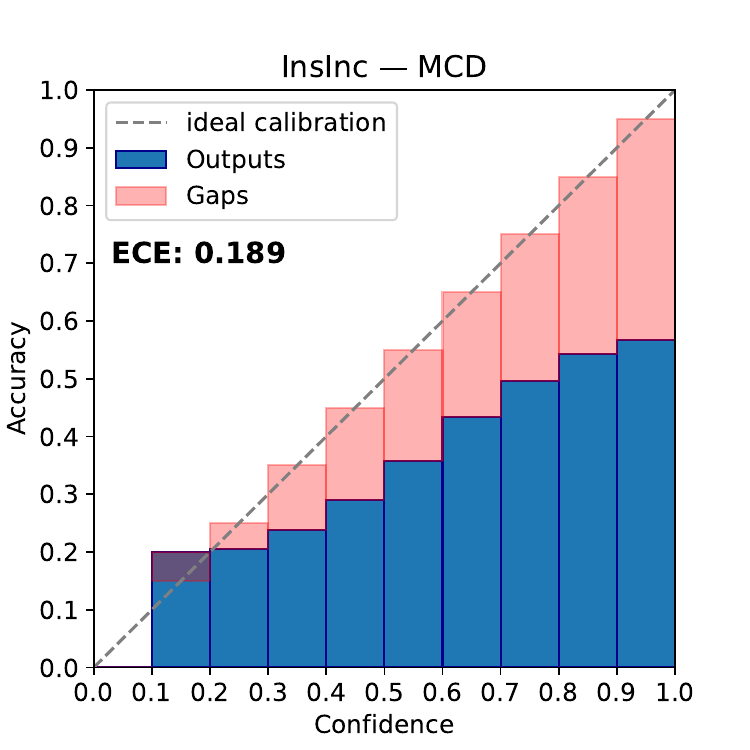}&\includegraphics[width=45mm]{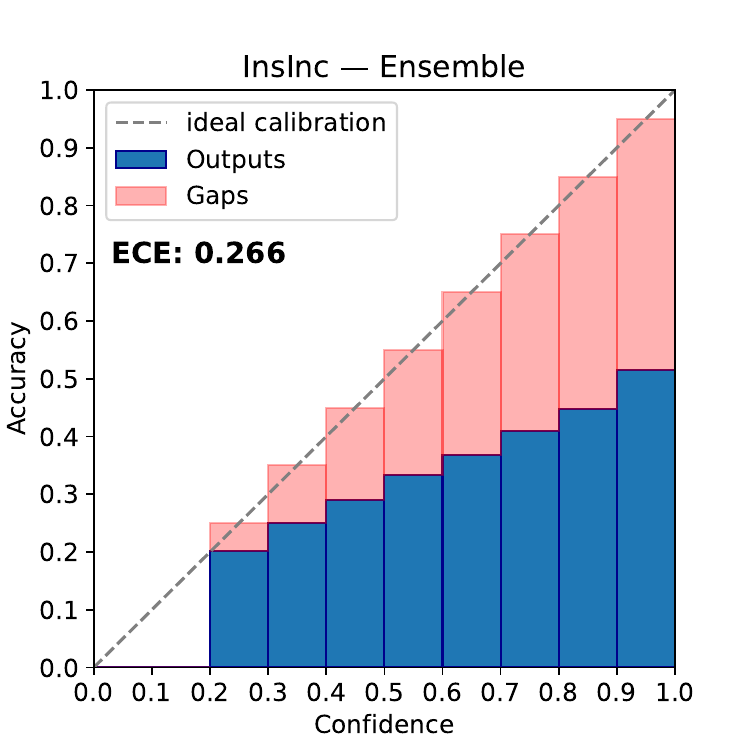}&\includegraphics[width=45mm]{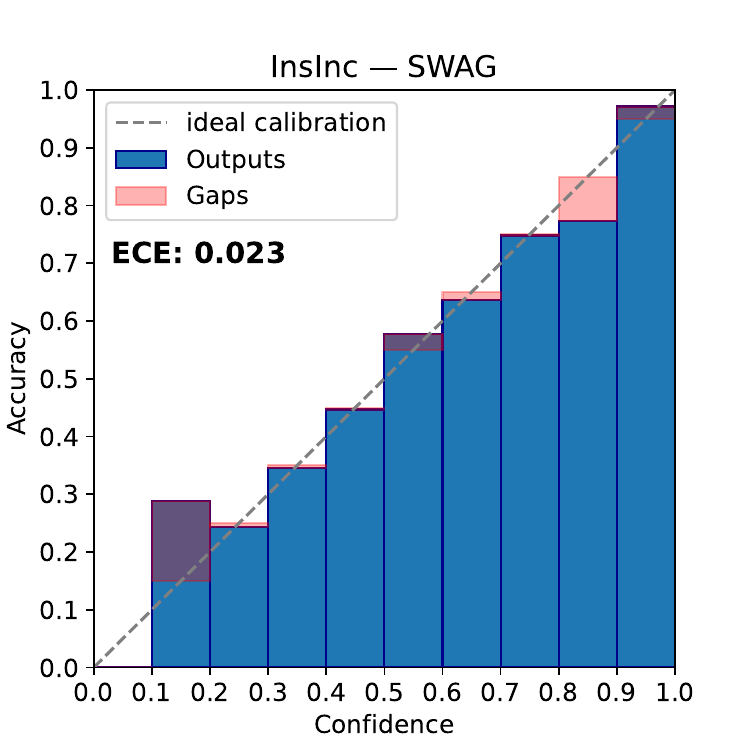}&\includegraphics[width=45mm]{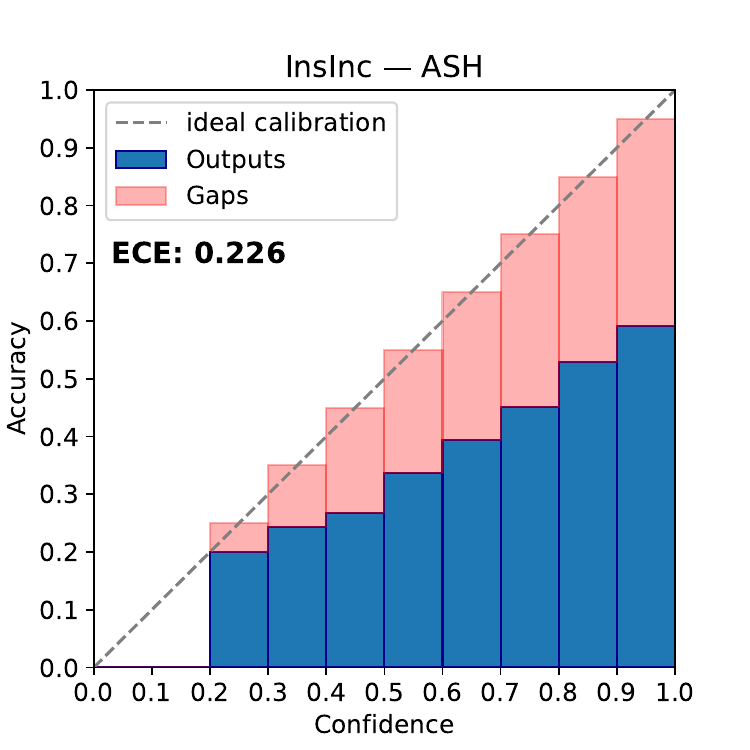}\\
         \hspace{-3mm}
          \includegraphics[width=45mm]{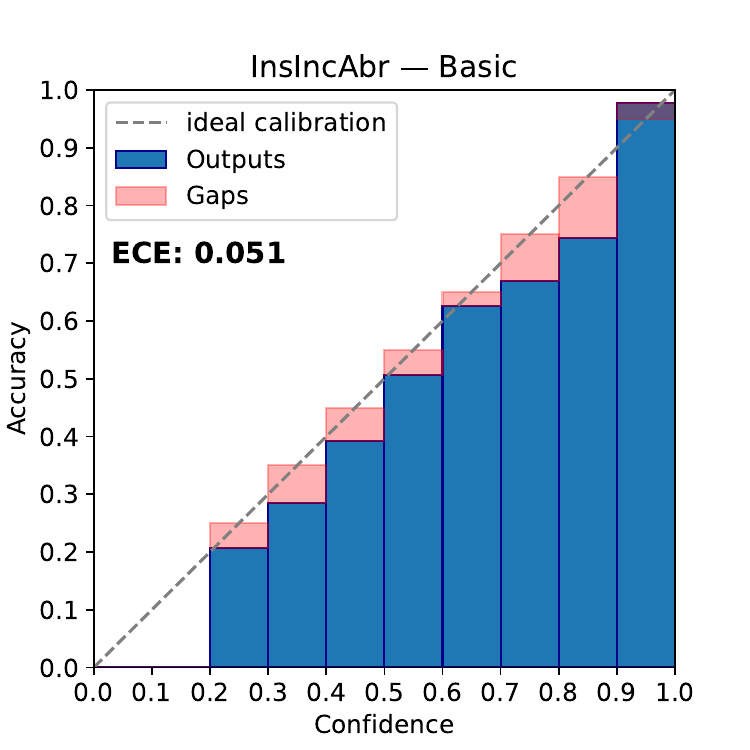}&\includegraphics[width=45mm]{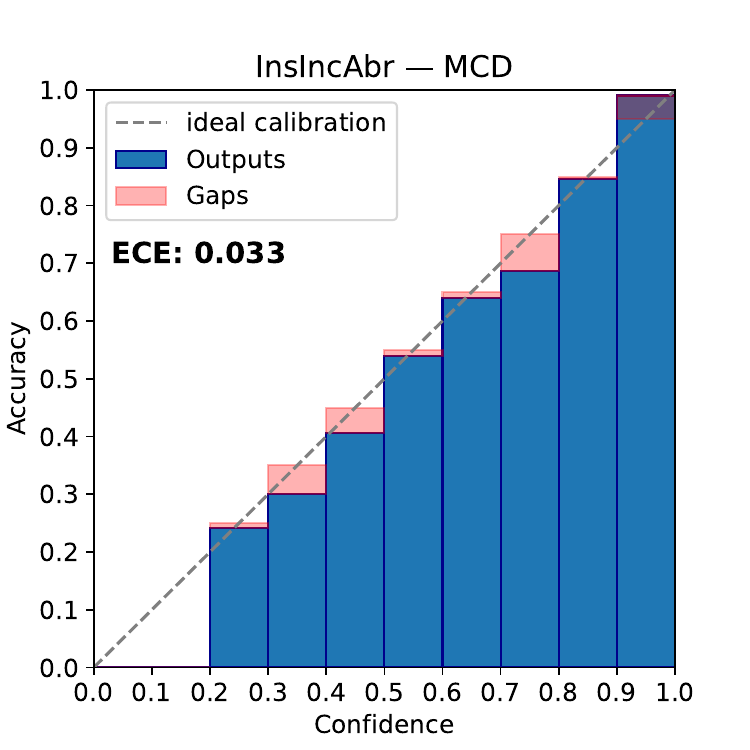}&\includegraphics[width=45mm]{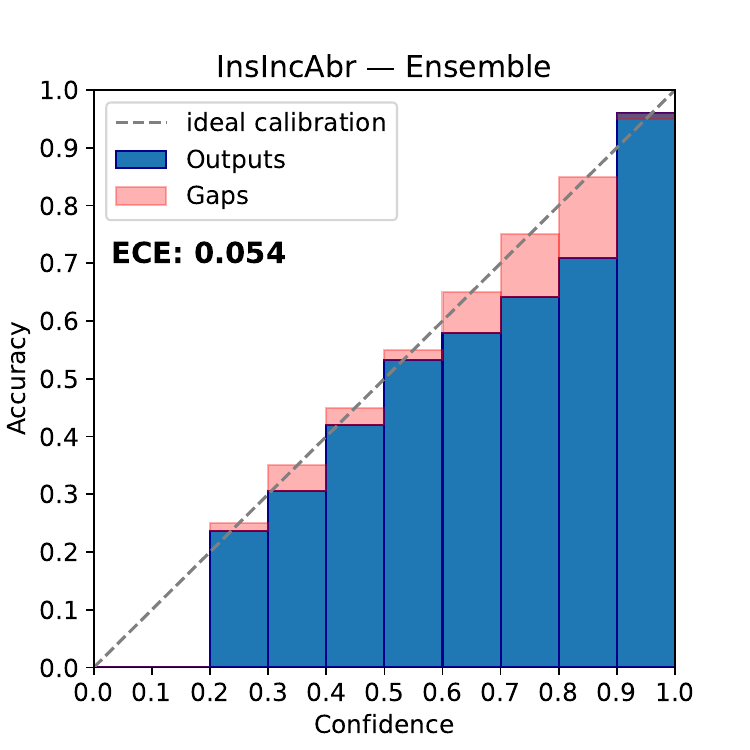}&\includegraphics[width=45mm]{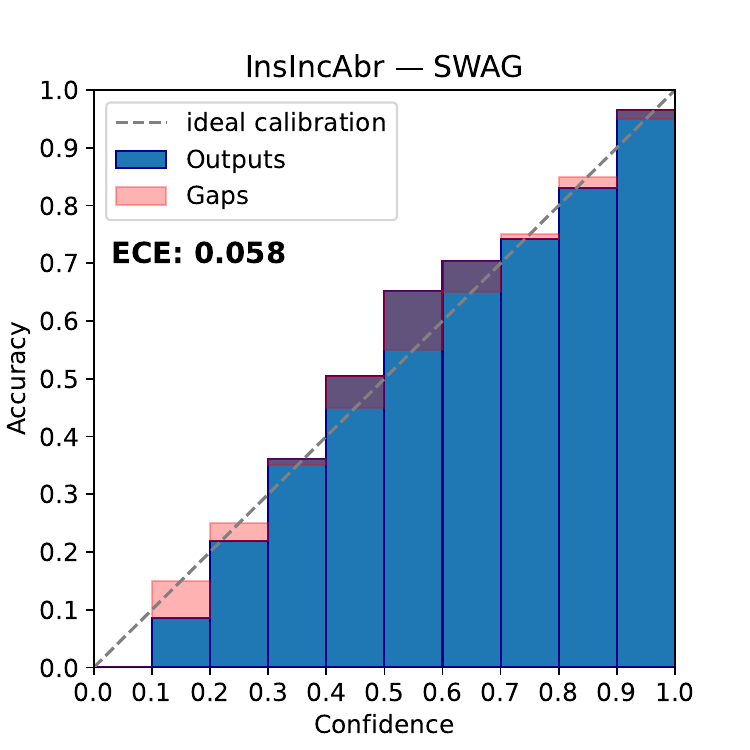}&\includegraphics[width=45mm]{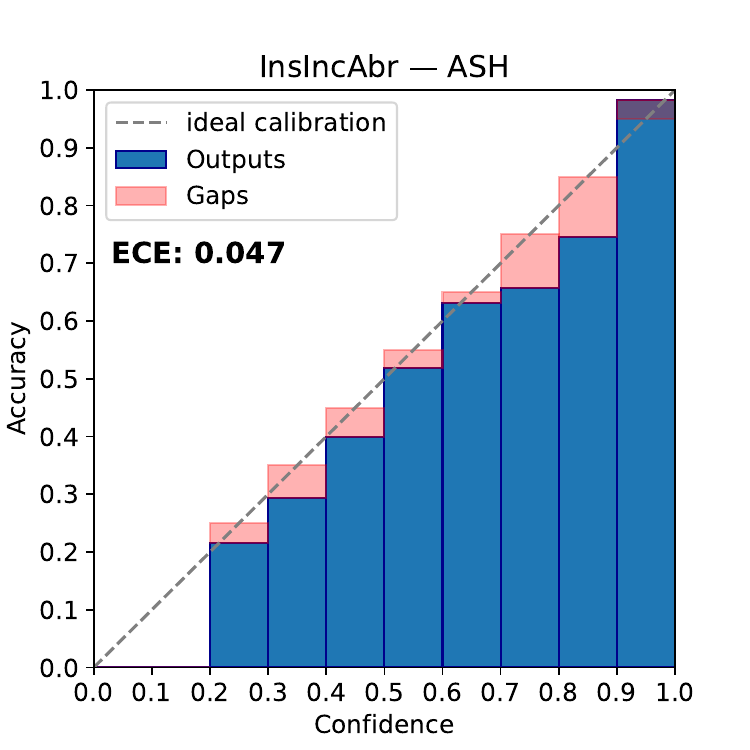}\\
          
      \end{tabular}
      \caption{Reliability diagrams of main experiment (2/3)}
      \label{fig:reliability_2}
  \end{sidewaysfigure}

\begin{sidewaysfigure}[!htbp]
      \centering
      \setlength{\tabcolsep}{-2pt}
      
   \vspace{10mm}
      \begin{tabular}{ccccc}
         \hspace{-3mm}
          \includegraphics[width=45mm]{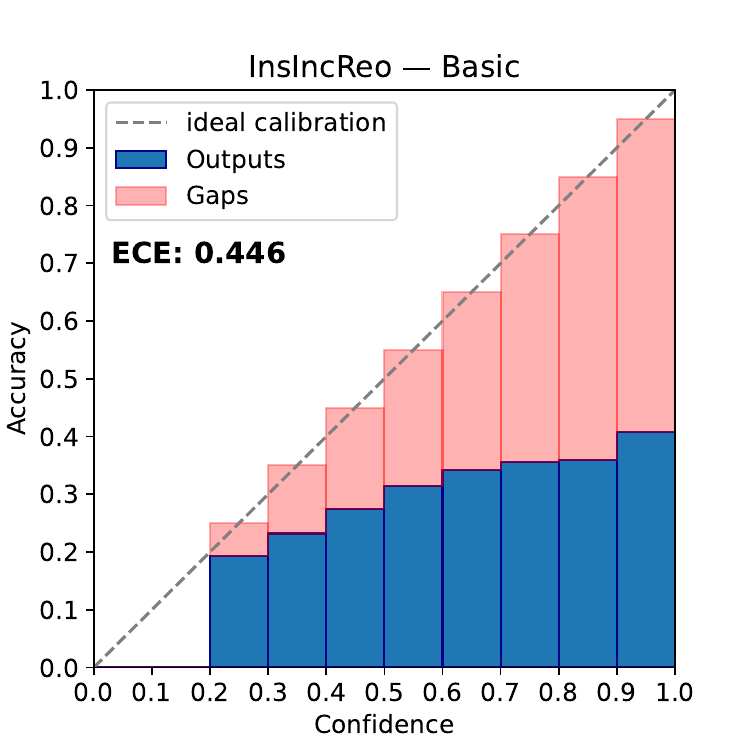}&\includegraphics[width=45mm]{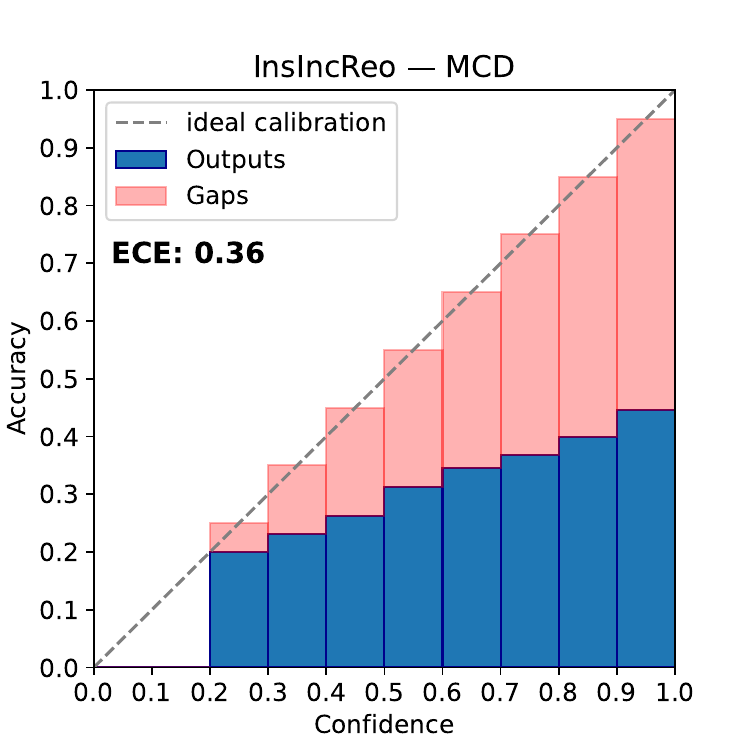}&\includegraphics[width=45mm]{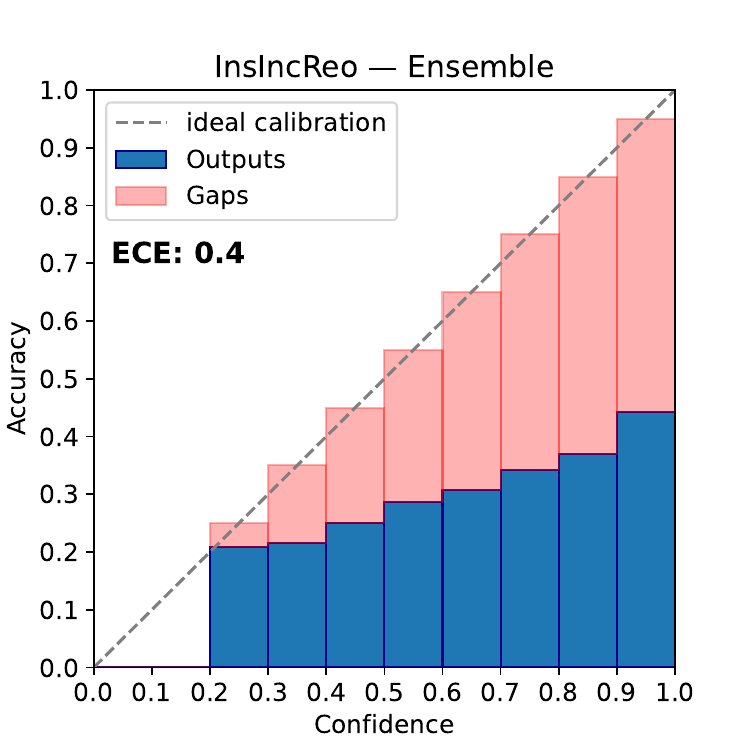}&\includegraphics[width=45mm]{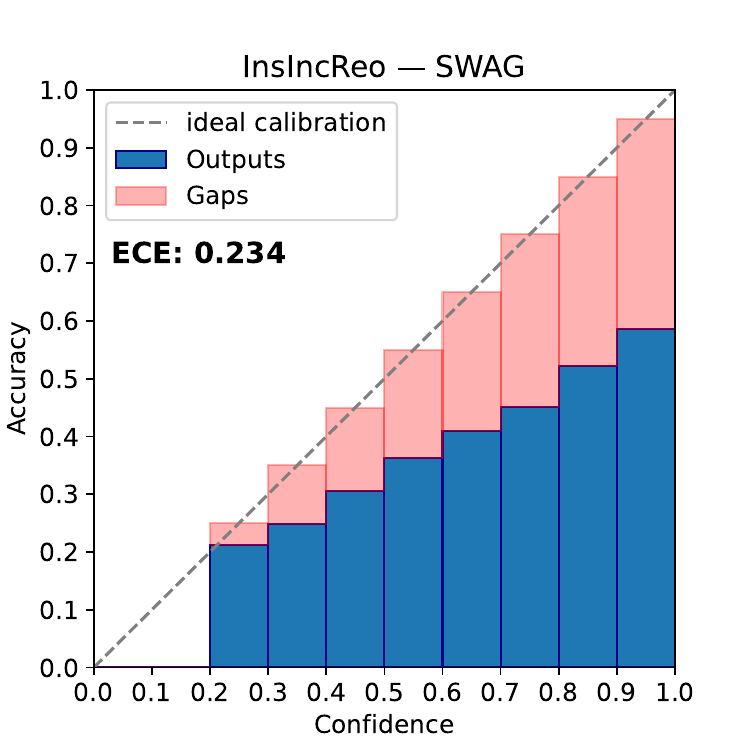}&\includegraphics[width=45mm]{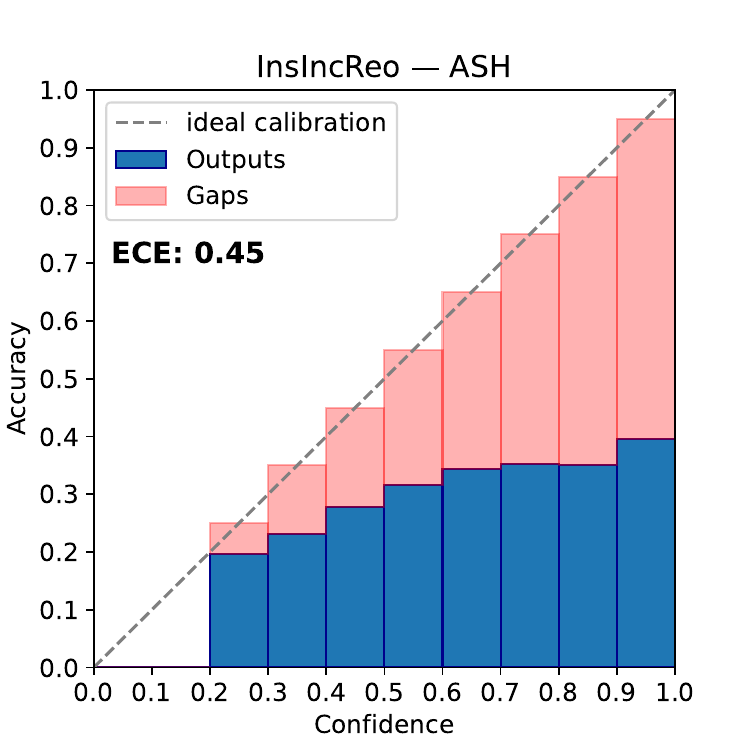}\\

      \end{tabular}
      \caption{Reliability diagrams of main experiment (3/3)}
      \label{fig:reliability_3}
  \end{sidewaysfigure}

We generated reliability diagrams ~\citep{GuoPSW17} in addition to Table \ref{table:final_comparison} and Figure \ref{fig:ECE} of the main experiment.
These diagrams illustrate the quantification of the ECE.
Hence, buckets of confidence values are compared to their average accuracy. 
Furthermore, the gaps to a perfect calibration are visualized.
Plots can be found in Figures \ref{fig:reliability_1} - \ref{fig:reliability_3}.
Consistent to Figure \ref{fig:ECE}, they show that the SWAG method offers the best calibration.\\
To validate the retraining positions found by the uncertainty based drift detection, we also conducted an experiment with equally and randomly distributed retraining positions.
We compared these against the SWAG-based drift detection.
Hence, the same amount of retrainings was triggered as found by the SWAG approach for each dataset (see Table \ref{table:equal_random_dist}).
While the retraining positions found by SWAGs uncertainty values yield significantly better predictions for \textit{Gas}, \textit{Electricity}, and \textit{InsAbr}, the opposite is the case for \textit{InsIncAbr} and \textit{InsIncReo}.
Here, the equally distributed approach for retrainings offers noticeably better results.
For \textit{Rialto} and \textit{InsInc} there are only slight differences between all three methods. 
Nevertheless, the detection based approach still offers the best overall performance.\\
As we repeated all of these experiments five times with different random seeds, we also include the standard deviations in Tables \ref{table:std_baseline} - \ref{table:std_val}.

\begin{table}[h!]
   \centering
   
   \caption{Retraining position validation.}
   \label{table:equal_random_dist}
   
    \begin{tabular}{cccc}
      \toprule
        & SWAG & Equal dist. & Random dist.\\
        \midrule
       Gas & \textbf{0.46} (52) & 0.387 (52) & 0.38 (52)\\
       \midrule
       Electricity & \textbf{0.419} (7) & 0.346 (7) & 0.351 (7)\\
       \midrule
       Rialto & 0.54 (52) & 0.555 (52) & \textbf{0.557} (52)\\
       \midrule
       InsAbr&  \textbf{0.514} (6) & 0.459 (6) & 0.484 (6)\\
       \midrule
       InsInc & \textbf{0.301} (4) & \textbf{0.301} (4) & 0.293 (4)\\
       \midrule
       InsIncAbr & 0.445 (25) &  \textbf{0.483} (25) & 443 (25)\\
       \midrule
       InsIncReo & 0.302 (21) & \textbf{0.344} (21) & 0.32 (21)\\
       \bottomrule
   \end{tabular}  
\end{table}
  
\begin{table}[h!]
    \centering
    
    \caption{Standard deviations of baseline experiment without retrainings.}
    \label{table:std_baseline}
    
        \begin{tabular}{cccccc}
        \toprule
        & Basic & MCD & Ensemble & SWAG & ASH\\
        \midrule
        Gas &0.0249    &  0.0437&      0.0119&      0.0274&      0.0404\\
        \midrule
        Electricity & 0.016  &     0.0093&      0.0125&      0.0235&      0.014\\
        \midrule
        Rialto & 0.0102    &  0.0118      &0.0054&      0.0035&      0.0014\\
        \midrule
        InsAbr & 0.0066  &    0.0017&      0.0017&      0.0014&      0.0022\\
        \midrule
        InsInc & 0.0099   &   0.0101&      0.0101&      0.0096&      0.0075\\
        \midrule
        InsIncAbr &0.0067    &  0.0035&      0.0042&      0.0086&      0.005\\
        \midrule
        InsIncReo  & 0.0073   &   0.0109&      0.004&       0.0033&      0.0059 \\
        \bottomrule
    \end{tabular}  
    
\end{table}

\begin{table}[h!]
    \centering
    
    \caption{Standard deviations of main experiment with ADWIN detection.}
    \label{table:std_main}
    
        \begin{tabular}{cccccc}
        \toprule
        & Basic & MCD & Ensemble & SWAG & ASH\\
        \midrule
        Gas & 0.0331   &   0.0438     & 0.0366&      0.0203      &0.0347\\
        \midrule
        Electricity & 0.0199    &  0.0368      &0.0356      &0.0187      &0.0136\\
        \midrule
        Rialto & 0.0034    &  0.0028      &0.0030&      0.0036&      0.0026 \\
        \midrule
        InsAbr & 0.0060  &    0.0082      &0.0107&      0.0251&      0.0101\\
        \midrule
        InsInc & 0.0156    &  0.0125      &0.0138&      0.0162&      0.0158\\
        \midrule
        InsIncAbr & 0.0021  &    0.0050&      0.0096&      0.0203&      0.0078\\
        \midrule
        InsIncReo  & 0.0088   &   0.0072&      0.0111&      0.0113&      0.0095 \\
        \bottomrule
    \end{tabular}  
    
\end{table}

\begin{table}[h!]
    \centering
    
       \caption{Standard deviations of retraining position validation experiment.}
    \label{table:std_val}
    
        \begin{tabular}{cccc}
        \toprule
        & SWAG & Equal dist. & Random dist.\\
        \midrule
        Gas & 0.0203  & 0.0094 & 0.0305\\
        \midrule
        Electricity&0.0187      &0.0208 & 0.0398\\
        \midrule
        Rialto & 0.0036&  0.0013  & 0.0047 \\
        \midrule
        InsAbr &   0.0251&   0.0039 & 0.0247  \\
        \midrule
        InsInc &    0.0172&    0.0021 &0.0316 \\
        \midrule
        InsIncAbr &    0.0203&   0.0081  & 0.0469\\
        \midrule
        InsIncReo  &    0.0113&   0.0036 & 0.0156\\
        \bottomrule
    \end{tabular}  
\end{table}


\pagebreak
\subsection{Hyperparameter Tuning}
\label{tuning}
To tune the hyperparameters, the main experiment was run for several configurations with the same seed. 
Tables in the following show the MCC based on all predictions and the number of retrainings in parentheses. 

For \textbf{MCD} the only additional hyperparameter is the number of stochastic forward passes $T$.
As Table \ref{table:mcd_tuning} reveals, we tested $T =  25,\;50,\;75\;\text{and} \;100$.
Although  $T=25$ had the best performance in the majority of our experiments, this is not really representative for the overall performance. 
In fact, discrepancies are rather slight in datasets where $T=25$ yields the best performance, while it is significantly outperformed in other datasets.
We found that  $T=100$ offers the most balanced performance across all datasets.
The additional computational cost is also negligible as $T=100$ triggers the least amout of retrainings and thus incurrs the lowest execution time for all experiments combined.

\begin{table}[h!]
  \centering
  
  \caption{MCC values of MCD with $25$, $50$, $75$, and $100$ forwards passes.}
  \label{table:mcd_tuning}
   \begin{tabular}{ccccc}
      \toprule
       & $T=25$ & $T=50$ & $T=75$ & $T=100$\\
      \midrule
      Gas & 0.418 (49) & 0.443 (48) & 0.41 (44)  & \textbf{0.451} (46) \\
      \midrule
      Electricity & 0.365 (8) & 0.386 (10)& 0.363 (11)& \textbf{0.415} (8)\\
      \midrule
      Rialto & 0.554 (59) & \textbf{0.56} (61)& 0.554 (59) & 0.553 (59)\\
      \midrule
      InsAbr & 0.509 (9)& 0.491 (6)& \textbf{0.521} (10)& 0.481 (5)\\
      \midrule
      InsInc & \textbf{0.218} (2) & 0.216 (1)& 0.217 (1)& 0.216 (1)\\
      \midrule
      InsIncAbr & \textbf{0.54} (25)& 0.538 (23)& 0.538 (23) & 0.538 (22)\\
      \midrule
      InsIncReo  & \textbf{0.249} (21)& 0.24 (19) & 0.24 (20) & 0.235 (18)\\
      \midrule
      Total exec. time & 4712s & 4712s & 4875s & \textbf{4638}s\\
      \bottomrule
  \end{tabular}  

\end{table}

Similar to MCD there is only one hyperparameter for the \textbf{Ensemble}.
Namely, the number of members $M$ which was set to three, five, and seven during our tests, as shown in Table \ref{table:ensemble_tuning}.
Here we found very slight differences overall.
Thus, we choose the version with the least computational cost, which is $M=3$.

\begin{table}[h!]
  \centering
  
  \caption{MCC values of an ensemble of $3$, $5$, and $7$ members.}
  \label{table:ensemble_tuning}
  
   \begin{tabular}{cccc}
      \toprule
       & $M=3$ & $M=5$ & $M=7$\\
       \midrule
      Gas & 0.479 (48) & \textbf{0.494} (51) & 0.479 (52)\\
      \midrule
      Electricity & 0.422 (10) & 0.407 (9)& \textbf{0.426} (10)\\
      \midrule
      Rialto & \textbf{0.529} (46) & \textbf{0.529} (48)& 0.526 (51)\\
      \midrule
      InsAbr & 0.474 (4)& \textbf{0.505} (8)& 0.494 (8)\\
      \midrule
      InsInc & \textbf{0.259} (3) & 0.194 (1)& 0.255 (2)\\
      \midrule
      InsIncAbr & \textbf{0.53} (24)& 0.514 (26)& 0.508 (22)\\
      \midrule
      InsIncReo  & 0.231 (21)& \textbf{0.255} (22) & 0.25 (18)\\
      \midrule
      Total exec. time & \textbf{12912}s & 19511s & 31092s\\
      \bottomrule
  \end{tabular}  
\end{table}

Other than the previous methods, \textbf{SWAG} comes with several hyperparameters. 
First, the influence of the number of weight samples $S$ drawn from the approximated distribution was tested. 
Therefore, the rank $K$ was set to $25$, and weights were updated every epoch starting at the first iteration. 
As shown by Table \ref{table:swag_tuning}, $S = 100$ offers the most balanced performance. 
The higher execution time compared to $S = 50$ and $S = 75$ is the result of more retrainings in datasets such as InsAbr and InsIncReo. 
Consequently, there is a noticeable performance gap in said datasets which mitigates the slower execution.
Onwards, the effect of the rank $K$ was studied with a fixed $S$. 
As seen in Table \ref{table:swag_tuning_K}, the initial rank of $K=25$ slightly outperformed the other settings.
Starting at later epochs and reducing the update frequency for the SWAG method showed no mentionable improvements neither in performance, nor in execution time. 
Thus the final configuration was $S = 100$ and $K = 25$ with updates in every epoch beginning at the start of training.

\begin{table}[h!]
    \centering
    
    \caption{MCC values of SWAG with $25$, $50$, $75$, and $100$ weight samples.}
    \label{table:swag_tuning}
    
     \begin{tabular}{ccccc}
        \toprule
         & $S=25$ & $S=50$ & $S=75$ & $S=100$\\
         \midrule
        Gas & 0.436 (49) & 0.433 (57) & \textbf{0.456} (55)  & 0.455 (53) \\
        \midrule
        Electricity & 0.414 (10) & \textbf{0.435} (11)& 0.343 (13)& 0.396 (10)\\
        \midrule
        Rialto & 0.544 (53) & \textbf{0.546} (54)& 0.543 (49) & 0.541 (53)\\
        \midrule
        InsAbr & \textbf{0.543} (9)& 0.503 (6)& 0.517 (7)& 0.542 (8)\\
        \midrule
        InsInc & 0.283 (3) & 0.29 (4)& \textbf{0.304} (4)& 0.296 (3)\\
        \midrule
        InsIncAbr & 0.51 (25)& 0.487 (23)& \textbf{0.528} (23) & 0.504 (22)\\
        \midrule
        InsIncReo  & 0.332 (31)& 0.282 (16) & 0.311 (20) & \textbf{0.335} (28)\\
        \midrule
        Total exec. time & 5694s & 5018s & \textbf{4913}s & 5514s\\
        \bottomrule
    \end{tabular}  
\end{table}

\begin{table}[h!]
    \centering
    
    \caption{MCC values of SWAG with $S=100$ and $K=10,\;25,$ and $40$.}
    \label{table:swag_tuning_K}
     \begin{tabular}{cccc}
        \toprule
         & $K=10$ & $K=25$ & $K=40$\\
         \midrule
        Gas & 0.443 (54) & \textbf{0.455} (53) & 0.435 (55)  \\
        \midrule
        Electricity & \textbf{0.412} (10) & 0.396 (10)& 0.392 (13)\\
        \midrule
        Rialto & 0.543 (54) & 0.541 (53)& \textbf{0.548} (50) \\
        \midrule
        InsAbr & 0.521 (7)& \textbf{0.542} (8)& 0.54 (8)\\
        \midrule
        InsInc & 0.255 (3) & 0.296 (3)& \textbf{0.318} (4)\\
        \midrule
        InsIncAbr & 0.487 (26)& 0.504 (22)& \textbf{0.514} (21)\\
        \midrule
        InsIncReo  & 0.316 (28)& \textbf{0.335} (28) & 0.289 (22) \\
        \midrule
        Total exec. time & 5603s & 5514s & \textbf{5472}s\\
        \bottomrule
    \end{tabular}  
\end{table}

All three \textbf{ASH} versions introduced by Djurisic et al.~\citep{djurisic2022extremely} were tested with pruning percentages between 60\% and 90\% in the penultimate layer. 
As Table \ref{table:ash_tuning} reveals, the best results performance was achieved by the ASH-p version with a rather low pruning percentage of 60\%.
This is surprising, as ASH-p was the worst method in tests from Djurisic et al. where it served as a baseline. 
Furthermore, experiments have shown that higher pruning percentages were hurting performance.
Lastly, the placement of the pruning layer was tested for the previous best configuration. 
While differences were slight, the best performance was reached when pruning in the penultimate hidden layer (i.e. third last overall layer) as seen in Table \ref{table:ash_tuning_layer}.

\begin{table}[h!]
    \centering
    
    \caption{MCC values of ASH-p (top), ASH-b (middle) and, ASH-s (bottom) for pruning percentages 60\%, 70\%, 80\% and 90\%.}
    \label{table:ash_tuning}
    
     \begin{tabular}{ccccc}
        \toprule
         & 60\% & 70\% & 80\% & 90\%\\
         \midrule
        \multirow{3}{*}{Gas} & \textbf{0.443} (42) & \textbf{0.443} (42) & 0.398 (47)  & 0.293 (60) \\
        & 0.407 (38) & 0.407 (38) & 0.357 (39) & 0.388 (48)\\
        & 0.397 (28) & 0.397 (28) & 0.396 (26) & 0.347 (25)\\
        \midrule
        \multirow{3}{*}{Electricity} & \textbf{0.475} (12) & 0.395 (10)& 0.408 (6)& 0.422 (13)\\
        & 0.347 (8) & 0.464 (10) & 0.444 (10) & 0.412 (12)\\
        & 0.318 (4) & 0.339 (3) & 0.338 (3) & 0.398 (4)\\
        \midrule
        \multirow{3}{*}{Rialto} & 0.539 (42) & 0.537 (41)& 0.526 (45) & 0.447 (43)\\
        &0.551 (38) & 0.56 (36) & 0.564 (39) & 0.442 (41)\\
        &0.569 (35) & 0.561 (35) & \textbf{0.575} (35) & 0.45 (39)\\
        \midrule
        \multirow{3}{*}{InsAbr} & \textbf{0.496} (7)& 0.474 (10)& 0.435 (7)& 0.322 (8)\\
        &0.471 (4) & 0.386 (9) & 0.426 (5) & 0.301 (5)\\
        &0.491 (6) & 0.405 (7) & 0.435 (5) & 0.342 (9)\\
        \midrule
        \multirow{3}{*}{InsInc}& 0.217 (1) & \textbf{0.252} (5)& 0.179 (2)& 0.172 (3)\\
        &0.237 (2) & 0.155 (3) & 0.162 (3) & 0.153 (2)\\
        &0.23 (2) & 0.196 (3) & 0.196 (5) & 0.223 (3)\\
        \midrule
        \multirow{3}{*}{InsIncAbr} & 0.502 (24)& 0.502 (24)& 0.477 (24) & 0.336 (22)\\
        &0.473 (25) & 0.473 (25) & 0.435 (19) & 0.424 (20)\\
        &\textbf{0.526} (17) & \textbf{0.526} (17) & 0.453 (21) & 0.419 (24)\\
        \midrule
        \multirow{3}{*}{InsIncReo}  & 0.232 (17)& \textbf{0.254} (17) & 0.208 (17) & 0.171 (21)\\
        &0.202 (12) & 0.196 (13) & 0.142 (8) & 0.168 (21)\\
        &0.214 (9) & 0.171 (6) & 0.223 (13) & 0.116 (17)\\
        \midrule
        \multirow{3}{*}{Total exec. time} & 3570s & 3551s & 3718s & 3657s\\
        &3034s & 3122s & 3026s & 3483s\\
        &2804s & \textbf{2715}s & 3240s & 3297s\\
        \bottomrule
    \end{tabular}  
\end{table}

\begin{table}[h!]
    \centering
    
    \caption{MCC values of ASH-p with a pruning percentage of 60\% at outputlayer - 1, -2, and -3.}
    \label{table:ash_tuning_layer}
    
        \begin{tabular}{cccc}
        \toprule
            & $L-1$ & $L-2$ & $L-3$\\
            \midrule
        Gas & \textbf{0.443} (42) & 0.441 (36) & 0.399 (39)  \\
        \midrule
        Electricity & \textbf{0.475} (12) & 0.423 (13)& 0.453 (13)\\
        \midrule
        Rialto & 0.539 (42) & \textbf{0.545} (43)& \textbf{0.545} (43) \\
        \midrule
        InsAbr & \textbf{0.496} (7)& 0.494 (6)& \textbf{0.496} (8)\\
        \midrule
        InsInc & 0.217 (1) & \textbf{0.237} (3)& 0.234 (3)\\
        \midrule
        InsIncAbr & 0.502 (24)& \textbf{0.526} (21)& 0.524 (24)\\
        \midrule
        InsIncReo  & 0.232 (17)& \textbf{0.259} (21) & 0.245 (20) \\
        \midrule
        Total exec. time & \textbf{3570}s & 3611s & 3646s\\
        \bottomrule
    \end{tabular}  
\end{table}
\end{document}